%% file: iclr2023_conference_tinypaper.tex
\documentclass{article} 
\usepackage{iclr2023_conference_tinypaper,times}

\input{math_commands.tex}

\usepackage{hyperref}
\usepackage{url}

\usepackage{xcolor}
\usepackage{soul}
\newcommand{\mathcolorbox}[2]{\colorbox{#1}{$\displaystyle #2$}}

\newtheorem{lemma}{Lemma}

        {\medskip}

        {\hspace*{\fill}$\Box$\par}
        {\hspace*{\fill}$\Box$\par\vspace{4mm}}
        {\hspace*{\fill}$\Box$\par}

        {\hspace*{\fill}$\Box$\par}

\title{Reward Bound for Behavioral Guarantee of Model-based Planning Agents}

\author{Zhiyu An, Xianzhong Ding \& Wan Du \\
Department of Computer Science and Engineering\\
University of California, Merced\\
Merced, CA, USA \\
\texttt{\{zan7, xding5, wdu3\}@ucmerced.edu} \\
}

\iclrfinalcopy 
\begin{document}

\maketitle

\begin{abstract}
Recent years have seen an emerging interest in the trustworthiness of machine learning-based agents in the wild, especially in robotics, to provide safety assurance for the industry. Obtaining behavioral guarantees for these agents remains an important problem. In this work, we focus on guaranteeing a model-based planning agent reaches a goal state within a specific future time step. We show that there exists a lower bound for the reward at the goal state, such that if the said reward is below that bound, it is impossible to obtain such a guarantee. By extension, we show how to enforce preferences over multiple goals.
\end{abstract}

\section{Introduction \& Preliminaries}

We have witnessed a rapid integration of machine learning-driven agents in our lives, such as industrial robots and autonomous driving vehicles. However, research on the trustworthiness of these agents lags behind \cite{landers2023deep, borg2018safely}. Behavioral Guarantee, an important technique to improve the reliability of autonomous agents, has been explored in this context \cite{kress2018synthesis}. While it is relatively straightforward to guarantee an agent's behavior to avoid imminent fault (for instance, adding distance sensors to prevent a robot's collision with people), it is more challenging to guarantee the same for faults that are over-the-horizon (for instance, guaranteeing an autonomous vehicle's behavior in traffic scenarios several seconds into the future). In this work, we study the problem of guaranteeing a model-based planning agent reaches a goal state within a future time step from a reward function-design perspective.

\textbf{Environment Formulation.} We consider problems formulated by a discrete-time Markov Decision Process (MDP) $\mathcal{M}:\{\mathcal{S}, \mathcal{A}, r, \mathcal{P}\}$ with state space $\mathcal{S}$, action space $\mathcal{A}$, reward function $r:\mathcal{S}\rightarrow\mathbb{R}$, dynamics function $\mathcal{P}(s'|s, a)$. At time step $t$, the agent is in state $s_t\in\mathcal{S}$, take some action $a_t\in\mathcal{A}$, transitions to the next state $s_{t+1}\sim \mathcal{P}(s_t, a_t)$ and receives reward $r_t=r(s_{t+1})$. The goal is to maximize $\sum_{t=0}^\infty r(s_t)$. For the sake of simplicity, we assume $r$ is non-negative.

\textbf{Agent Definition.} Model-based Reinforcement Learning (MBRL) \cite{wang2019benchmarking} is widely used to solve MDP in system control \cite{an2023clue}, robotics \cite{mendonca2023alan}, etc. It approximates the optimal policy using two components: a dynamics model and an optimizer. The dynamics model $\hat{\mathcal{P}}$ learns the discrete-time dynamics of the environment from historical data $\{(s_t, a_t, s_{t+1})\}_N$. Then, $\hat{\mathcal{P}}$ is used by the optimizer to select an appropriate action by solving $a[:] = \arg\max_{a[:]}\sum_{t=0}^H\gamma^tr(\hat{\mathcal{P}}(s_t, a_t))$, where $a[:] = [a_0,\cdots, a_{H-1}]$ and $\gamma\in[0, 1]$ is a discount factor. The agent then takes the first action in the sequence, $a[0]$, and re-plan in the next time step. To solve for this action sequence, an optimizer usually generates $a[:]$ according to some rule, predicts the state trajectory $s[:] = \bigcup_{a_t\in a[:]} \mathcal{P}(s_t, a_t)$, and evaluates the total reward of trajectories $\sum_{t=0}^H r(s_t), s_t\in s[:]$ \cite{botev2013cross, rao2009survey, jorge2006numerical}.

\textbf{Problem Definition.} Assuming the agent is currently in $s_0$, we consider the problem of guaranteeing the agent reaches a goal state $s_g$ within the next $J$ steps with a probability of $1$.

\section{Main Result}

Note that if the underlying MDP is not deterministic, such a guarantee is impossible. For the sake of the analysis, we assume that the underlying MDP is deterministic and the agent's model $\hat{\mathcal{P}}$ resembles the true dynamics $f$ with sufficient accuracy to not change the evaluation of rewards, i.e., $r(\hat{\mathcal{P}}(s_t, a_t)) = r(\mathcal{P}(s_t, a_t))$ for any $s_t, a_t$. We propose two necessary conditions to obtain such a guarantee. First, the agent must have enough time to transition to the goal state. Formally, we construct the bootstrapped forward reachable set \cite{liu2021algorithms} of the agent from $s_0$, shown in Eq. \ref{eq: forward reachable set}. The first necessary condition is that $s_g\in R(s_0, J)$.
\begin{equation}\label{eq: forward reachable set}
    R(s_0, H) = \bigcup_{t=0}^{H-1} \{s_{t+1} | s_{t+1} = \mathcal{P}(s_t, a_t); \forall a\in\mathcal{A}\}
\end{equation}
Second, the trajectories containing the goal must have a total reward larger than any other trajectories. This condition is necessary for the trajectories containing the goal to be selected by the optimizer, thus leading the agent to transition towards the goal. Formally, let the set of trajectories starting from $s_0$ and containing the goal be $\{g[:]\}$, it is necessary for Eq. \ref{eq: condtion 2} to be true.
\begin{equation}\label{eq: condtion 2}
    \exists g[:], \;\sum_{s_t\in g[:]}\gamma^tr(s_t) > \sum_{s_t\in s[:]}\gamma^tr(s_t) \;\;\forall s[:]\in\{s[:] \;|\; s[0] = s_0,\; s[:] \cup \{s_g\} = \emptyset\}
\end{equation}

\lemma Given that the underlying MDP is deterministic and the model is accurate, if Eq. \ref{eq: condtion 2} is false, then the probability of the agent reaching $s_g$ within $J$ steps is strictly less than $1$.\normalfont\label{lemma 1}

The proof of Lemma \ref{lemma 1} is postponed to Appendix \ref{app, Proof of lemma 1}. If the agent is deterministic and is able to fully explore the entire forward reachable set to find the max-reward trajectory, then the above two conditions are the sufficient conditions for behavioral guarantee. Now, we consider two application scenarios of these conditions: single goal state, and multiple goal states with preference order.

\subsection{Single Goal State}\label{Sec: single goal state}

In this case, the agent is expected to reach a single $s_g$. This is often the final state of a task, and the agent stops after reaching this state. In practice, $r(s_g)$ is usually set to an arbitrary large number, such as $9999$. This approach is prone to fault, e.g. when other states are assigned a similarly large number, which will misguide the agent to reach other states instead. We show that:
\begin{equation}\label{eq: single goal state}
\gamma^Jr(s_g) > \sum_{s_t\in s[:]} \gamma^tr(s_t)    \;\;\forall s[:]\in\{s[:] \;|\; s[0] = s_0,\; s[:] \cup \{s_g\} = \emptyset\}
\end{equation}
is sufficient to guarantee the agent reaches the goal. The proof is postponed to Appendix \ref{app, Proof of Section 2.1}.

\subsection{Multiple Goal States with Preference Order}

In this scenario, we have multiple goal states ordered by our preference $s_g^1 \succ s_g^2 \cdots \succ s_g^N$. If the forward reachable set contains multiple goal states, the reward function should guarantee that the agent reaches the goal state with highest preference, ignoring other goal states.

\lemma Given that the underlying MDP is deterministic and the model is accurate, if the least preferred goal $s_g^N$ satisfies Eq. \ref{eq: single goal state} and all other goals satisfy:
\begin{equation}\label{eq. lemma 2}
    s_g^i \succ s_g^j \iff \gamma^Jr(s_g^i) > \sum_{k = 0}^{J-1} \gamma^Jr(s_g^j)
\end{equation}
, then the agent reaches the highest preference goal in a forward reachable set with probability $1$.\normalfont\label{lemma 2}

The proof is postponed to Appendix \ref{app, Proof of Lemma 2}.

\section{Conclusion}


We proposed necessary and sufficient conditions to guarantee a model-based planning agent's success in reaching a goal within a future time step. By extension, we showed that the proposed conditions can be applied to enforce preferences over multiple goals.

\subsubsection*{Acknowledgements}
This work was supported in part by NSF Grant \#2239458 and UC National Laboratory Fees Research Program Grant \#69763. Any opinions, findings, and conclusions expressed in this material are those of the authors and do not necessarily reflect the views of the funding agencies.

\subsubsection*{URM Statement}
The authors acknowledge that all authors of this work meets the URM criteria of ICLR 2024 Tiny Papers Track.

\bibliography{iclr2023_conference_tinypaper}
\bibliographystyle{iclr2023_conference_tinypaper}

\appendix
\section{Appendix}

\subsection{Proof of Lemma \ref{lemma 1}}\label{app, Proof of lemma 1}

Assume the agent is deterministic and is able to explore the entire forward reachable set. Eq. \ref{eq: condtion 2} is false indicates that there exists some $s[:]$ such that 
\begin{equation}
\sum_{s_t\in s[:]} \gamma^tr(s_t) \ge \sum_{s_t\in g[:]} \gamma^tr(s_t),\;\;\forall g[:] 
\end{equation}
In the best case, we have $\sum_{s_t\in s[:]} \gamma^tr(s_t) = \sum_{s_t\in g[:]} \gamma^tr(s_t)$, and there is only one such $s[:]$ and many $g[:]$. If there are many trajectories with equal total rewards, the optimizer chooses one of these trajectories at random, and the probability of eventually selecting all actions that actuates trajectory $s[:]$ is larger than $0$. In other cases, there will be a trajectory $s[:]$ whose total reward is larger than any $g[:]$. Thus $g[:]$ has a $0$ probability to be selected.

Otherwise, if the agent employs stochastic optimization (e.g. random shooting \cite{rao2009survey}), the statement in the subject is automatically true. \hfill$\Box$

\subsection{Proof of Section \ref{Sec: single goal state}}\label{app, Proof of Section 2.1}

The first condition holds trivially. Since $r$ is non-negative and $s_g \in g[:]$, 
\begin{equation}
    \begin{aligned}
        \sum_{s_t\in g[:]}\gamma^tr(s_t) &> r(s_g)\\
        &> \gamma^Jr(s_g) \\
        &> \sum_{s_t\in s[:]} \gamma^tr(s_t) \\
        \sum_{s_t\in g[:]}\gamma^tr(s_t) &> \sum_{s_t\in s[:]}\gamma^tr(s_t)
    \end{aligned}
\end{equation}
for $\forall s[:]\in\{s[:] \;|\; s[0] = s_0,\; s[:] \cup \{s_g\} = \emptyset\}$, which satisfies Eq. \ref{eq: condtion 2} and hence the second condition. \hfill$\Box$

\subsection{Proof of Lemma \ref{lemma 2}}\label{app, Proof of Lemma 2}

First, we check that a trajectory that contains any goal state will be selected over trajectories without goal state. To simplify the notation, we use $g^i[:]$ to denote the trajectory that contains goal $s_g^i$ and does not contain any goal with higher preference.

By the definition of Lemma \ref{lemma 2}, we have $s_g^N$ satisfies Eq. \ref{eq: single goal state}. Hence:
\begin{equation}
    \sum_{k = 0}^{J-1} \gamma^Jr(s_g^N) \ge \sum_{s_t\in g^N[:]}\gamma^tr(s_t) > \sum_{s_t\in s[:]}\gamma^tr(s_t) \;\;\forall s[:]\in\{s[:] \;|\; s[0] = s_0,\; s[:] \cup \{s_g\} = \emptyset\}
\end{equation}
which shows that trajectories containing $s_g^N$ will be selected over trajectories without goal state. By Eq. \ref{eq. lemma 2}, 
\begin{equation}\label{eq. appendix proof of lemma 2}
    \sum_{s_t\in g^i[:]}\gamma^tr(s_t) \ge \gamma^Jr(s_g^i) > \sum_{s_t\in s[:]}\gamma^tr(s_t) \;\;\forall s[:]\in\{s[:] \;|\; s[0] = s_0,\; s[:] \cup \{s_g\} = \emptyset\}
\end{equation}
for any $i$ where $s_g^i \succ s_g^N$. Since $s_g^N$ has the least preference, Eq. \ref{eq. appendix proof of lemma 2} is true for all goal states other than $s_g^N$. Therefore, trajectories containing any goal state will be selected over trajectories without goal state.

Next, we check that trajectories containing the highest preference goals will be selected over other trajectories. By Eq. \ref{eq. lemma 2}, for any $i, j$ where $s_g^i \succ s_g^j$,
\begin{equation}
    \begin{aligned}
        \mathcolorbox{yellow}{\sum_{s_t\in g^i[:]}\gamma^tr(s_t)} \ge \gamma^Jr(s_g^i) &> \sum_{k = 0}^{J-1} \gamma^Jr(s_g^j) > \mathcolorbox{yellow}{\sum_{s_t\in g^j[:]}\gamma^tr(s_t)}\\
        \mathcolorbox{yellow}{\sum_{s_t\in g^i[:]}\gamma^tr(s_t)} &> \mathcolorbox{yellow}{\sum_{s_t\in g^j[:]}\gamma^tr(s_t)}
    \end{aligned}
\end{equation}
Which proves that the total reward of the trajectory that contains higher preference goals will be strictly greater than the trajectories that contains lower preference goals. Hence, the agent will always select the trajectory with higher preference goals. \hfill$\Box$

\end{document}

%% file: math_commands.tex
\usepackage{amsmath,amsfonts,bm}

\def\eqref#1{equation~\ref{#1}}

\def\1{\bm{1}}

\DeclareMathAlphabet{\mathsfit}{\encodingdefault}{\sfdefault}{m}{sl}
\SetMathAlphabet{\mathsfit}{bold}{\encodingdefault}{\sfdefault}{bx}{n}

